\documentclass[letterpaper, 10 pt, conference]{ieeeconf}  % Comment this line out if you need a4paper

\IEEEoverridecommandlockouts                              % This command is only needed if 
\usepackage{amsmath} % assumes amsmath package installed
\usepackage{amssymb}  % assumes amsmath package installed
\usepackage{xcolor}
\usepackage{svg}
\usepackage{bm}
\usepackage{optidef}
\usepackage{url}

\newtheorem{assumption}{Assumption}

\newtheorem{definition}{Definition}
\title{\LARGE \bf
An optimal open-loop strategy for handling a flexible beam\\with a robot manipulator}

\author{Shamil Mamedov$^{1,2}$, Alejandro Astudillo$^{1,2}$, Daniele Ronzani$^{1,2}$,\\Wilm Decré$^{1,2}$, Jean-Philippe Noël$^{1}$, Jan Swevers$^{1,2}$% <-this % stops a space
\thanks{This research was supported by the FWO-Vlaanderen through SBO project ELYSA for cobot applications (S001821N).}% <-this % stops a space
\thanks{$^1$The MECO Research Team, KU Leuven, 3000 Leuven, Belgium. }
\thanks{$^2$The DMMS Lab, Flanders Make, 3001 Leuven, Belgium.}%
}

\begin{document}

\maketitle
\thispagestyle{empty}
\pagestyle{empty}

%%%%%%%%%%%%%%%%%%%%%%%%%%%%%%%%%%%%%%%%%%%%%%%%%%%%%%%%%%%%%%%%%%%%%%%%%%%%%%%%
\begin{abstract}
Fast and safe manipulation of flexible objects with a robot manipulator necessitates measures to cope with vibrations. Existing approaches either increase the task execution time or require complex models and/or additional instrumentation to measure vibrations. This paper develops a model-based method that overcomes these limitations. It relies on a simple pendulum-like model for modeling the beam, open-loop optimal control for suppressing vibrations, and does not require any exteroceptive sensors.
We experimentally show that the proposed method drastically reduces residual vibrations -- at least 90\% -- and outperforms the commonly used input shaping (IS) for the same execution time. Besides, our method can also execute the task faster than IS with a minor reduction in vibration suppression performance. The proposed method facilitates the development of new solutions to a wide range of tasks that involve dynamic manipulation of flexible objects.
\end{abstract}

%%%%%%%%%%%%%%%%%%%%%%%%%%%%%%%%%%%%%%%%%%%%%%%%%%%%%%%%%%%%%%%%%%%%%%%%%%%%%%%%
\section{INTRODUCTION}
Many industries extensively use flexible materials \cite{Saadat2002IndustrialApplications}.
Naive handling of flexible objects with a robot arm may introduce large vibrations. Existing feedback solutions \cite{Kapsalas2018ARXbeam, Liu2009Adaptive} require accurate sensing of the vibrations using an additional sensor and complex analytical or data-driven models. On the other hand, existing feedforward solutions increase the task execution time \cite{Singer1990ishaping}.
Therefore, the industry can substantially benefit from new methods for fast handling of flexible objects that are strong in performance and simple in implementation.

This paper investigates the general problem of manipulating a flexible beam with a rigid robot arm \cite{Kapsalas2018ARXbeam}. The problem involves modeling, parameter estimation, control, and perception. We assume a structured industrial environment and do not address the perception. For sensing vibrations of the beam, we do not use exteroceptive sensors -- such as external force-torque sensors at the end-effector or a camera -- only a joint torque sensor/estimator. This constraint on instrumentation increases the problem's complexity and the practical value of the developed solutions for economic reasons. 

Any model-based control method requires a model of the system. 
Beams are infinite dimensional systems; they are accurately modeled by partial differential equations that are computationally demanding to solve and are seldom used in control and trajectory planning. In robotics, for accurate modeling of flexible beams, researchers make simplifying assumptions, e.g., separability of spatial and time modes as in the assumed mode method \cite{Sakawa1985}, or apply discretization methods such as the finite element method \cite{Zhou2002NonlinearIsh, Liu2009Adaptive}. Another related approach is flexible multibody dynamics in relative (to the rigid body mode) \cite{Hoshyari2019Disney} or absolute nodal coordinates \cite{Myhre2014ANC}. The model parameters in the above-mentioned methods are often obtained from CAD models because, in practice, it is difficult to estimate them.
\begin{figure}
    \centering
    \includegraphics[keepaspectratio, width=0.6\linewidth]{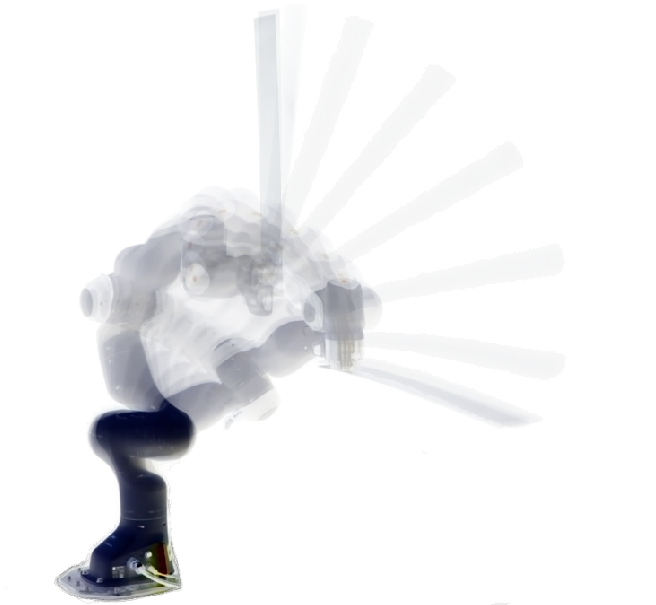}
    \caption{A stroboscopic photo of the Franka Panda handling a flexible beam}
    \vspace{-0.35cm}
    \label{fig:cover_photo}
\end{figure}
Data-driven methods (system identification) approach modeling beam dynamics differently; they infer the model structure from data \cite{Kapsalas2018ARXbeam}. 

In this paper, we use a simple lumped parameter model for the beam modeling that only considers the first natural frequency. The model is computationally fast compared with more accurate models and physically interpretable, unlike purely data-driven methods \cite{Kapsalas2018ARXbeam}. Parameters of the model can be estimated from data or analytically from material properties. 

The most crucial aspect of flexible object handling lies in vibrations suppression.  
Input shaping (IS) is one of the most well-established techniques for suppressing vibrations of linear systems \cite{Singer1990ishaping}. Despite its simplicity, input shapers modify the original trajectory and extend the motion time. In robotics, modification of the joint trajectories yields changes in the end-effector trajectory. To avoid end-effector trajectory changes, IS can be applied to the normalized arc length  \cite{Zhao2016ZeroTimeDelayZV} or the operational space trajectory \cite{Aribowo2015ZV2lin}. To counteract increased motion time, \cite{Zhao2016ZeroTimeDelayZV} proposes accelerating the original motion by the amount of delay. 
Zhout et al. \cite{Zhou2002NonlinearIsh} developed a nonlinear IS for suppressing flexible payload vibrations. Instead of shaping joint accelerations, the authors shaped modal excitation forces. However, retrieving joint velocities or accelerations from shaped modal forces is not trivial and sometimes not unique.

Neither linear nor nonlinear input shapers can handle input constraints; for example, a feasible trajectory in the joint space after being shaped in the operational space might become infeasible. In contrast, optimal control can explicitly consider state and input constraints.  
In \cite{Reinhold2019DynSloshing}, the authors used direct time-optimal control for the sloshing-free transport of liquids with a robot arm. They formulated the optimal control problem (OCP) in the operational space, with the decision variables being the end-effector translational accelerations.
% Authors could achieve sloshing free transport both in simulation and experimentally. 
Indirect optimal control \cite{Pappalardo2018adjoint, Hoshyari2019Disney} and model predictive control (MPC) \cite{Rupert2015MPCvsIS, Katliar2018NMPC, Boscariol2010FlexMPC} have also been used for vibration suppression. MPC solves an underlying OCP at every sampling instant and is naturally more robust to model errors. However, it comes at a high computation cost and complexity since MPC requires online/real-time measurements and computation. To reduce the computational burden, researchers often drastically reduce the horizon of the MPC or linearize the nonlinear dynamics \cite{Boscariol2010FlexMPC, Rupert2015MPCvsIS} rendering MPC conservative. 

We approach vibration suppression from a direct optimal control point of view similar to \cite{Reinhold2019DynSloshing}. However, instead of formulating an OCP in the operational space, we do it in the joint space where constraints can be handled more naturally, and arm dynamics can be fully exploited. In terms of control strategy, we opt for an open-loop formulation instead of feedback control, as feedback control requires real-time estimation of the beam vibrations, which are challenging to obtain without exteroceptive sensors. Because of the underlying optimization problem, our approach can be computationally slow compared with solutions that leverage IS. To address it, we propose an efficient numerical implementation that substantially reduces computation time compared with a naive implementation. 

To briefly summarize, our contributions are:
\begin{itemize}
    \item a model that is simple yet effectively captures the complex dynamics of the system;
    \item an OCP for handling the beam that drastically reduces residual vibrations and allows to trade-off vibrations suppression and task execution time;
    % \item an efficient numerical formulation of the OCP;
    \item experimental validation of the proposed method and comparison with IS.
\end{itemize}

This paper is organized as follows: Section \ref{sec:modeling} addresses the modeling of the robot arm and the beam, Section \ref{sec:param_est} introduces two methods for estimating the parameters of the beam model. In Section \ref{sec:method}, we discuss the proposed control method; in Section \ref{sec:experiments}, we present experimental results, followed by a discussion. Section \ref{sec:disc_conc} concludes the paper and provides directions for future research.

\section{MODELING} \label{sec:modeling}
\begin{figure}[t]
    \centering
    % \includesvg[width=\linewidth]{figures/modeling_assump.svg}
    \includegraphics[keepaspectratio, width=\linewidth]{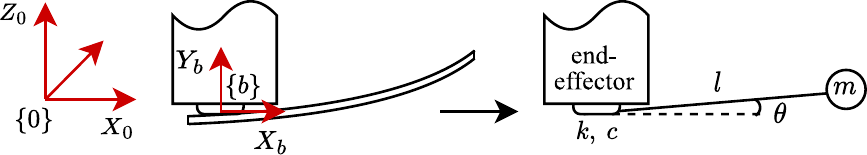}
    \caption{Schematic representation of approximating a beam attached to the end-effector of a robot arm with a simple pendulum of length $l$ and a lumped mass $m$ attached to the end-effector by means of a passive revolute joint with stiffness $k$ and damping $c$}
    \vspace{-0.35cm}
    \label{fig:modeling_assump}
\end{figure}

In this section, we develop a model of the setup and discuss underlying assumptions.

\subsection{Arm dynamics}
For a robot arm with $n_\mathrm{dof}$ degrees of freedom ($\mathrm{dof}$), let $\boldsymbol q \in \mathbb{R}^{n_\mathrm{dof}}$ be the vector of joint positions and assume that: \begin{assumption}
% You can also put an optional title as: \begin{assumption}[Title]
\label{as:arm_model}
The internal joint controller can accurately track the joint reference trajectories.
\end{assumption}
Then, a double integrator model accurately describes the arm's dynamics: 
\begin{align}  \label{eq:kin_model}
    \ddot{\boldsymbol q} = \boldsymbol u,
\end{align}
where $\ddot{\boldsymbol q} \in \mathbb{R}^{n_\mathrm{dof}}$ is the vector of joint accelerations, and $\boldsymbol u \in \mathbb{R}^{n_\mathrm{dof}}$ is the vector of inputs (reference joint accelerations).

\subsection{Beam dynamics}
For modeling the beam dynamics, we make another critical and simplifying assumption: 
\begin{assumption}
\label{as:beam_model}
The beam can be approximated by a simple pendulum connected to the end-effector of an arm through passive revolute joint with stiffness $k$ and damping $c$, as shown in Fig. \ref{fig:modeling_assump}. 
\end{assumption}
By making such assumption, we consider only the first natural frequency of a beam and only the lateral vibrations.

To derive the pendulum dynamics using the Lagrange formulation \cite[Ch. 7]{sciavicco2001book}, 
% we need to describe mass position and velocity using kinematics of the robot. 
let $\bm p_{m}^0 \in \mathbb{R}^{3}$ denote the position of the pendulum mass $m$ in the robot's base frame
\begin{align} \label{eq:mass_pos}
    \bm p_{m}^0 = \bm p_{b}^{0} + l \bm R_{b}^{0} \bm R_z(\theta) {\bm i},
    %
    % \dot{\bm p}_m = \dot{\bm p} + l \dot \theta \bm R \frac{d \bm R_z(\theta)}{d \theta} \overrightarrow{\bm i} + l \dot{\bm R} \bm R_z(\theta) \overrightarrow{\bm i} 
\end{align}
where $\bm p_{b}^{0}$ is the position of the origin of the frame $\{b\}$ in the base frame $\{0\}$,  $\bm R_{b}^{0} \in \mathrm{SO(3)}$ is the transformation from frame $\{b\}$ to the base frame $\{0\}$, $\bm R_z(\theta) \in \mathrm{SO(3)}$ is a rotation matrix around $Z_b$ axis, $\theta$ is the angular position of the pendulum and ${\bm i} = [1\ 0\ 0]^\top$ is a unit vector. From now on, we drop superscript ${}^0$ for convenience. Differentiating \eqref{eq:mass_pos} yields the velocity of the pendulum mass
\begin{align} \label{eq:mass_vel}
    \dot{\bm p}_m = \dot{\bm p}_b + l \dot \theta \bm R_b \frac{d \bm R_z(\theta)}{d \theta} {\bm i} + l \dot{\bm R_b} \bm R_z(\theta) {\bm i}. 
\end{align}
Given the expressions for the position and velocity of the pendulum mass we formulate the kinetic $K$, potential $P$, and dissipation $D$ functions necessary to derive the dynamic equation:
\begin{align}
    K = \frac{1}{2} \dot{\bm p}_m^\top \dot{\bm p}_m, \ P = \frac{1}{2} k \theta^2 - m \bm g^\top \bm p_m,\ D = \frac{1}{2} c \dot \theta^2,
\end{align}
where $\bm g = [0\ 0\ -9.81]^\top\ \mathrm{m}/\mathrm{s}^2$  is the gravity acceleration vector. Using the Lagrange equations
\begin{align}
    \frac{d}{dt}\left(\frac{\partial \mathcal{L}}{\partial \dot \theta} \right) - \frac{\partial \mathcal{L}}{\partial \theta} = - \frac{\partial D}{\partial \dot \theta},
\end{align}
with $\mathcal{L} = K - P$ being the Lagrangian, 
and properties of the rotation matrices, we obtain the final expression for the pendulum dynamics
\begin{align}
    \ddot \theta = &- 2 \zeta \omega_n \dot \theta - \omega_n^2 \theta + \frac{1}{l}{\bm i}^\top \frac{d \bm R_z(\theta)}{d \theta}^\top \bm R_b^\top (\bm g - \ddot{\bm p}_{b}) \nonumber \\
    & - {\bm i}^\top \frac{d \bm R_z(\theta)}{d \theta}^\top \bm R_b^\top \bm S(\dot{\bm \omega}_b) \bm R_b \bm R_z(\theta) {\bm i} \label{eq:pend_dynamics} \\
    & + {\bm i}^\top\frac{d \bm R_z(\theta)}{d \theta}^\top \bm R_{b}^\top \bm S(\bm \omega_b)^\top \bm S(\bm \omega_b) \bm R_{b} \bm R_z(\theta){\bm i} \nonumber, 
\end{align}
where $\omega_{n} = \sqrt{k/(ml^2)}$, $\zeta = c/(2m\omega_n)$ are the natural undamped frequency and the damping ratio of the pendulum, respectively,  $\bm \omega_b \in \mathbb{R}^3$ and $\dot{\bm \omega}_b \in \mathbb{R}^3$ are the angular velocity and acceleration of the frame $\{b\}$ with respect to $\{0\}$ expressed in $\{0\}$ respectively and $\bm S(\cdot) \in \mathbb{R}^{3\times 3}$ is a skew-symmetric matrix. 

\subsection{Pendulum dynamics analysis}
Consider three different orientations of the $\{b \}$ frame: ($\mathcal{O}_1$) gravity acts laterally on the beam, ($\mathcal{O}_2$) gravity compresses the beam, and ($\mathcal{O}_3$) gravity extends the beam (see Fig. \ref{fig:eigen_rot}).

For each case, the dynamics of the pendulum alone in the absence of the external
excitation caused by a robot arm is:
\begin{align}
    &(\mathcal{O}_1):\quad \ddot \theta = -2 \zeta \omega_n \dot \theta - \omega_n^2 \theta + g_z \cos(\theta) /l \label{eq:O_neutral} \\
    &(\mathcal{O}_2): \quad \ddot \theta = -2 \zeta \omega_n \dot \theta - \omega_n^2 \theta - g_z \sin(\theta) /l \label{eq:O_up}\\
    &(\mathcal{O}_3):\quad \ddot \theta = -2 \zeta \omega_n \dot \theta - \omega_n^2 \theta + g_z \sin(\theta) /l \label{eq:O_down}
\end{align}
where $g_z=-9.81$ is the third component of the gravity vector $\bm g$. 
Eigenvalue analysis of the linearized models \eqref{eq:O_neutral}, \eqref{eq:O_up} and \eqref{eq:O_down} around $\bar \theta = 0$ shows that compared to ($\mathcal{O}_1$), for which $\omega_{n, \mathcal{O}_1} = \omega_n$, in ($\mathcal{O}_2$), gravity reduces the natural frequency $\omega_{n, \mathcal{O}_2} = \sqrt{\omega_{n}^2 + g_z/l}$. In contrast, in ($\mathcal{O}_3$), gravity increases the natural frequency $\omega_{n, \mathcal{O}_3} = \sqrt{\omega_{n}^2 - g_z/l}$. 
The demonstrated property of \eqref{eq:pend_dynamics} corresponds to the real beam dynamics as reported in \cite{Virgin2007} and experimentally shown in Section \ref{sec:experiments}. Thus, \eqref{eq:pend_dynamics} can be used for designing complex trajectories that involve changes in the position and orientation.

\subsection{Setup dynamics}
The setup dynamics combines arm and beam dynamics and is used later in the paper to develop the vibration suppression scheme. Let
$\bm x = \left[\bm q^\top\ \theta\ \dot{\bm q}^\top\ \dot \theta\right]^\top \in \mathbb{R}^{n_x}$ where $n_x = 2(n_{\mathrm{dof}}+1)$ denote the state of the system; then, the system dynamics can be expressed as:
\begin{align} \label{eq:setup_dynamics}
\dot{\bm x} = \bm f(\bm x,\bm u) =
\begin{bmatrix}
    \dot{\bm q}^\top &
    \dot \theta &
    \bm u^\top &
    % \text{rhs of \eqref{eq:pend_dynamics}}
    \ddot \theta
\end{bmatrix}^\top
\end{align}
where $\ddot \theta$ corresponds to the pendulum dynamics in \eqref{eq:pend_dynamics}.

\section{PARAMETER ESTIMATION} \label{sec:param_est}
In this section, we discuss two methods to estimate the parameters of a beam: a data-driven method and an analytical method in case experiments for parameter estimation are too costly or impossible to conduct.
\subsection{Data-driven method}
Explicit estimation of model parameters using system identification techniques requires reliable measurement of both system’s inputs and outputs. Inputs of \eqref{eq:pend_dynamics} –- accelerations and velocities of frame $\{b \}$ –- have to be calculated from joint acceleration estimates, which are noisy. The output is the component of the external wrench at the end-effector. For beam handling without the use of exteroceptive sensors, the output has to be estimated based on the joint torque sensor and the dynamic model of the robot. From our experience such estimate is not accurate enough for system identification. Thus, we employ a simple method that does not require reliable measurements or estimates of both inputs and outputs.

The applied data-driven approach originates from the theory of mechanical vibrations; namely, the damping ratio and the natural frequency of an underdamped second-order linear system can be estimated from the position and amplitude of $n$ peaks of its impulse response \cite{meirovitch2001fundamentals}. The parameter estimation procedure consists of:
\begin{enumerate}
    \item estimating the periods of damped oscillations and log-decrement
    \begin{align}
        T = \frac{1}{n}\sum_{k=1}^{n}\frac{t_k - t_0}{k},\ \delta = \frac{1}{n}\sum_{k=1}^{n} \frac{1}{k} \log \left(\frac{\hat \tau_z^b(t_0)}{\hat \tau_z^b(t_{k})} \right), \nonumber 
    \end{align}
    where $t_k$ is the location of the $k$-th peak and $\hat \tau_z^b(t_k)$ is its amplitude;
    \item estimating the damping ratio and the natural frequency 
    \begin{equation} \label{eq:zeta}
        \hat \zeta = \frac{\delta}{\sqrt{4\pi^2 + \delta^2}},\ \hat \omega_n = \frac{\sqrt{4\pi^2 + \delta^2}}{T}. \nonumber
    \end{equation}
\end{enumerate}
Ideally, the positions and amplitudes of two peaks are sufficient for estimating $\omega_n$ and  $\zeta$. However, we average over several peaks to mitigate measurement/estimation errors.                                            

Experiment design plays an essential role in parameter estimation. In our problem setting, there are two options for optimal excitation: applying an impulse to the beam while the arm is standing still or performing a step excitation in the operational space. Unfortunately, many robots do not accept discontinuous trajectories; therefore, step-like trajectories with continuous first and second derivatives are the best option to excite the beam dynamics.

\begin{figure}[]
    \centering
    % \includesvg[width=\linewidth]{figures/eigen_analysis.svg}
    \includegraphics[keepaspectratio, width=\linewidth]{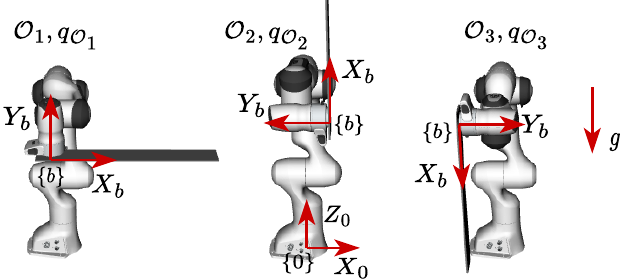}
    \vspace{-0.6cm}
    \caption{Orientations $\mathcal{O}_1,\ \mathcal{O}_2,\ \mathcal{O}_3$ of the frame $\{b\}$ and the corresponding joint positions $\bm q_{\mathcal{O}_1},\ \bm q_{\mathcal{O}_2},\ \bm q_{\mathcal{O}_3}$ used throughout the paper}
    \vspace{-0.35cm}
    \label{fig:eigen_rot}
\end{figure}

\subsection{Analytical method}
We apply the method developed in \cite{Sakawa1985} and \cite{Ge1997NewLumping}, where the spring-damper element is located at the pendulum mass. In this work, however, we locate the spring-damper element in the joint instead. Consider the Euler-Bernoulli beam equation with clamped-free boundary conditions. Let the deflection $w(t,x_b)$  be represented by only the first -- the most dominant -- separable harmonic mode: $w(x_b,t) = \phi(x_b) \xi(t)$, where $x_b$ is a spatial coordinate along $X_b$ (see Fig. \ref{fig:modeling_assump}), $\phi(x_b)$ is the first space-dependent mode shape function, and $\xi(t)$ is the first time-dependent harmonic mode. 

Based on the equivalence of kinetic and potential energies of the simple pendulum model and the Euler-Bernoulli beam equation, we obtain the following expressions for the pendulum mass and the spring constant
\begin{align}
    m_a = \frac{\rho }{l^2} \int_0^l \phi^2(x) dx, \ k_a = EI \int_0^l \left[ \frac{\partial \phi(x)}{\partial x} \right]^2 dx \nonumber, 
\end{align}
where $\rho$ is the uniform mass per unit length and $EI$ is the uniform flexural rigidity. The analytical method is limited; it does not give any estimate of the damping ration $\zeta$. Therefore, engineers have to take the value for $\zeta$ from catalogs or use their expert knowledge.

\section{OPTIMAL BEAM HANDLING} \label{sec:method}
We approach the beam handling by formulating it as an open-loop OCP. There are two reasons for such a choice. The first reason is constraint handling: to perform fast motions, the arm has to operate at its physical limits; thus, to avoid violating velocity and acceleration limits, we explicitly consider those constraints in the OCP formulation. The second reason is predictive control of vibrations (in open-loop): the arm needs to damp oscillations at the end, not during the whole motion, as it leads to overly conservative motions.

\subsection{Optimal control problem formulation}
Given an initial rest position of the setup $\bm q_{0}$ and $\theta_{0}$, the optimal controller has to drive the robot to a final rest pose $\bm p_{b, t_f}$, $\bm R_{b, t_f}$, and the corresponding equilibrium of the pendulum $\theta_{t_f}$ in a given time $t_f$ while minimizing a cost function and respecting constraints. 
Mathematically, the OCP can be formulated as follows:

%%%%%%%%%%%%%%%%%%%%%%%%%%%%%%%%%%%%%%%%%%%%%%%%%%%%%%%%%%%%%%%%%%%%%%%%%
\begin{mini!}|s|[2]
{ % Decision variables
\bm x, \bm{u}
}
{ % Objective function 
\int_{0}^{t_f} \left(||\bm{x-x}_0||_{\bm{W_x}} + ||\bm u||_{\bm{W_u}} + \rho ||\dot{\bm u}||_2 \right) dt   \label{eq:ocp_objective}
}
{\label{eq:ocp}}{}
% Constraints
\addConstraint{\dot{\bm x} = \bm f(\bm x, \bm u)}{\label{eq:ocp_constraint1}}{}
\addConstraint{\bm x(0) = [\bm q_0^\top\ \theta_{0}\ \bm 0^\top]^\top,\ \bm u(0) = \bm 0}{\label{eq:ocp_constraint2}}{}
\addConstraint{\bm p_{b}\left(\bm q(t_f)\right) = \bm p_{b, t_f},\ \dot{\bm q}(t_f) = \bm 0,\ \bm u(t_f) = \bm 0}{\label{eq:ocp_constraint3}}{}
\addConstraint{\theta(t_f) = \theta_{t_f},\ \dot \theta(t_f) = 0}{\label{eq:ocp_constraint4}}{}
\addConstraint{\bm e_O\left(\bm R_{b}\left(\bm q(t_f)\right), \bm R_{b, t_f}\right) = \bm 0_{3\times 1}}{\label{eq:ocp_constraint5}}{}
\addConstraint{\bm x \in \mathcal{X},\ \bm u \in \mathcal{U}, \ \dot{\bm u} \in \mathcal{J},}{\label{eq:ocp_constraint6}}{}
% End constraints
\end{mini!}
%
%%%%%%%%%%%%%%%%%%%%%%%%%%%%%%%%%%%%%%%%%%%%%%%%%%%%%%%%%%%%%%%%%%%%%%%%%
%
where $\bm{W_x} \in \mathbb{R}^{n_x \times n_x}, \bm{W_u} \in \mathbb{R}^{n_{\mathrm{dof}} \times n_{\mathrm{dof}}}$, and $\rho$ are the weights for penalizing deviation of states from the initial state, inputs, and jerks, respectively; $\bm e_O(\cdot) \in \mathbb{R}^3$ is a function for computing the orientation error between two frames \cite[Ch.3]{sciavicco2001book}; $\mathcal{X}$, $\mathcal{U}$, and $\mathcal{J}$ are feasible sets for states, controls, and rate of change of controls. In the problem formulation, we constrain controls to zero at time $0$ and $t_f$ to avoid discontinuous accelerations and infinite jerks at those points.
Later in the paper, we use a variation of \eqref{eq:ocp} as a baseline controller for beam handling and to generate excitation trajectories for the beam parameter estimation. For consistency, we define it here as:
\begin{definition}
\label{def:aOCP}
Arm OCP (aOCP) is an OCP similar to \eqref{eq:ocp} where the beam dynamics is neglected: \eqref{eq:ocp_constraint1} is replaced by the arm's dynamics \eqref{as:arm_model}, consequently \eqref{eq:ocp_constraint4} is eliminated, and the dimension of $\bm W_x$ is adjusted accordingly.
\end{definition}

\subsection{Implementation and efficient solution of the OCP}

In this work, we use CasADi \cite{Andersson2019casadi} to formulate \textit{aOCP} and OCP \eqref{eq:ocp} as nonlinear programs (NLP) following the multiple-shooting method with $N \in \mathbb{Z}_{>0}$ discretization points. We discretize \eqref{eq:ocp_constraint1} using a $4$th-order Runge-Kutta integrator. Derivatives of the control inputs are computed by using the forward Euler scheme ($\dot{\bm u}_k t_s = \bm{u}_{k+1} - \bm{u}_{k}$). The equilibrium positions of the spring $\theta_{0}$ and $\theta_{t_f}$ are computed from \eqref{eq:pend_dynamics} for a given rest configuration.

The NLPs are solved using the nonlinear optimization solver IPOPT \cite{wachter2006ipopt}, which implements an interior-point method and requires, among others, the evaluation of the Hessian of the Lagrangian of the NLPs, the evaluation of the constraints and their Jacobian, and the solution of a linear system involving the Karush-Kuhn-Tucker conditions of the NLPs, which are computationally expensive steps. Such high computational burden can be reduced by
% (i) use an efficient linear solver MA57 \cite{HSL} inside IPOPT, (ii) replace the exact Hessian of the Lagrangian with the L-BFGS Hessian approximation \cite{Liu_1989}, (iii) parallelize the evaluation of the discretized dynamics — allowed by the use of independent variables $(\bm{x}_k,\bm{u}_k)$ per discretization point in the multiple-shooting method \cite{Astudillo_par}—, and (iv) use just-in-time compilation to accelerate the evaluation of functions in the NLP and their derivatives.
\begin{itemize}
    \item using an efficient linear solver in IPOPT,
    \item replacing the exact Hessian of the Lagrangian with the L-BFGS Hessian approximation \cite{Liu_1989},
    \item parallelizing the evaluation of the discretized dynamics, allowed by the use of independent variables $(\bm{x}_k,\bm{u}_k)$ per discretization point,
    \item using efficient formulations of the robot kinematics and their derivatives,
    \item using just-in-time compilation of the functions in the NLPs.
\end{itemize}
The aforementioned techniques are used in this work to reduce the total solution time of OCP \eqref{eq:ocp}. Specifically, we use MA57 \cite{HSL} as linear solver and instruct the parallelization of the evaluation of \eqref{eq:ocp_constraint1} and their derivatives by using OpenMP via CasADi's \textit{map} construct as in \cite{Astudillo_par}. Moreover, we retrieve the computations of velocities and accelerations of the end-effector --i.e, first- and second-order kinematics required in the solution of \textit{aOCP} and OCP \eqref{eq:ocp}-- from the 
the forward pass of the recursive Newton-Euler algorithm, which exploits the sparsity of the kinematic model \cite{carpentier2018analytical} unlike algorithmic differentiation. Such efficient functions for kinematics (and their derivatives) are generated by using Pinocchio \cite{carpentier2019pinocchio}. The code used in this work is publicly available on a GitHub repository\footnote{\url{https://github.com/shamilmamedov/beam_handling}}.
%
% We retrieve the computations of velocities and accelerations of the end-effector --i.e, first- and second-order kinematics required in the solution of \textit{aOCP} and OCP \eqref{eq:ocp}-- from the 
% % The solution of \textit{aOCP} and OCP \eqref{eq:ocp} requires the computation of up to second-order kinematics of the end-effector. Instead of differentiating zero-order forward kinematics to compute velocities and accelerations of the end-effector, we retrieve such computations from 
% the forward pass of the recursive Newton-Euler algorithm, which exploits the sparsity of the kinematic model \cite{carpentier2018analytical} unlike algorithmic differentiation. Such efficient functions for kinematics (and their derivatives) are generated by using Pinocchio \cite{carpentier2019pinocchio}.
% , and are subject to common subexpression elimination to eliminate redundant atomic operations.
% and are required to compute not only the pose of the end-effector, but also the velocity and acceleration of the beam in both \textit{aOCP} and OCP \eqref{eq:ocp}. 

% ...
% The direct multiple-shooting method allows the parallel evaluation of the discretized dynamics constraints. 
%
% Show bottlenecks in the solution (Table with computation times)
%
% Now with L-BFGS, parallelization of multiple-shooting constraints does not contribute much to computation time reduction.
%
% Briefly mention common subexpression elimination
%
% Efficient linear solver in IPOPT (MA57 or Pardiso)
%

\section{EXPERIMENTS} \label{sec:experiments}
In this section, we present the experimental validation of the proposed approach. First, we address the parameter estimation; then, for three different tasks, we compare the performance of several vibration suppression controllers.

\subsection{Setup description}
Fig. \ref{fig:cover_photo} shows the setup used to validate our approach. It consists of a 7-$\mathrm{dof}$ Franka Emika Panda manipulator and a flexible beam with dimensions $60\times6\times1 \ \mathrm{mm}$  rigidly attached to the arm's end-effector. 
% This way, we simplify the problem (neglecting the first part of the problem: grasping the beam). 
The beam is made of stainless steel 316L with $\rho = 0.6296\ \mathrm{kg}/\mathrm{m}$, $EI=1.26667\ \mathrm{N}\cdot \mathrm{m}^2$.  
Inputs of the setup are reference joint velocities $\dot{\boldsymbol q}_r(t)$, while outputs are joint positions $\boldsymbol q(t)$ and velocities $\dot{\boldsymbol q}(t)$ as well as estimated filtered external torques $\hat{\boldsymbol \tau}_{\text{ext}}(t)$. The external wrench $\hat{\boldsymbol F} = [\hat F_x\ \hat F_y\ \hat F_z\ \hat \tau_x\ \hat \tau_y\ \hat \tau_z]^\top$ is estimated from filtered torque estimates provided by the robot's software, and  the manipulator Jacobian $\boldsymbol J_{b}$ as $\boldsymbol J_{b}(\boldsymbol q)^\top \hat{\boldsymbol \tau}_{\text{ext}} = \hat{\boldsymbol F}$.

\subsection{Parameter estimation}
We estimated the pendulum parameters using offline data-driven and analytical methods for comparison 
For the data-driven method, experiments consisted of three 1-dimensional trajectories of $0.15\ \mathrm{m}$, $0.25\ \mathrm{m}$, $0.35\ \mathrm{m}$ amplitude starting from three different orientations of the $\{b\}$ frame — $\mathcal{O}_1$, $\mathcal{O}_2$, and $\mathcal{O}_3$ as in Fig. \ref{fig:eigen_rot} — and moving along $Z_0$, $X_0$, and $-X_0$ axes, respectively. 

For each experiment, reference trajectories $\dot{\boldsymbol q}_r(t)$ were designed using aOCP; this way, we could design a fast step-like continuous trajectory to excite the beam dynamics and, at the same time, respect all the limits of the robot. We performed each motion six times to analyze the repeatability and reduce measurement noise by averaging. The outputs $\boldsymbol q(t)$, $\dot{\boldsymbol q}(t)$, and $\hat{\boldsymbol \tau}_{\text{ext}}(t)$ were logged at a frequency of $1$ kHz during the motion execution and 10 seconds afterward to record residual vibrations. Data processing of the collected data consisted of averaging $\hat{\boldsymbol \tau}_{\text{ext}}(t)$ over all experiments and computing the external wrench $\hat{\boldsymbol F} (t)$.  

\begin{figure}%[ht]
    \centering
    % \includesvg[width=\linewidth]{figures/Ti_deltai.svg}
    \includegraphics[width=\linewidth]{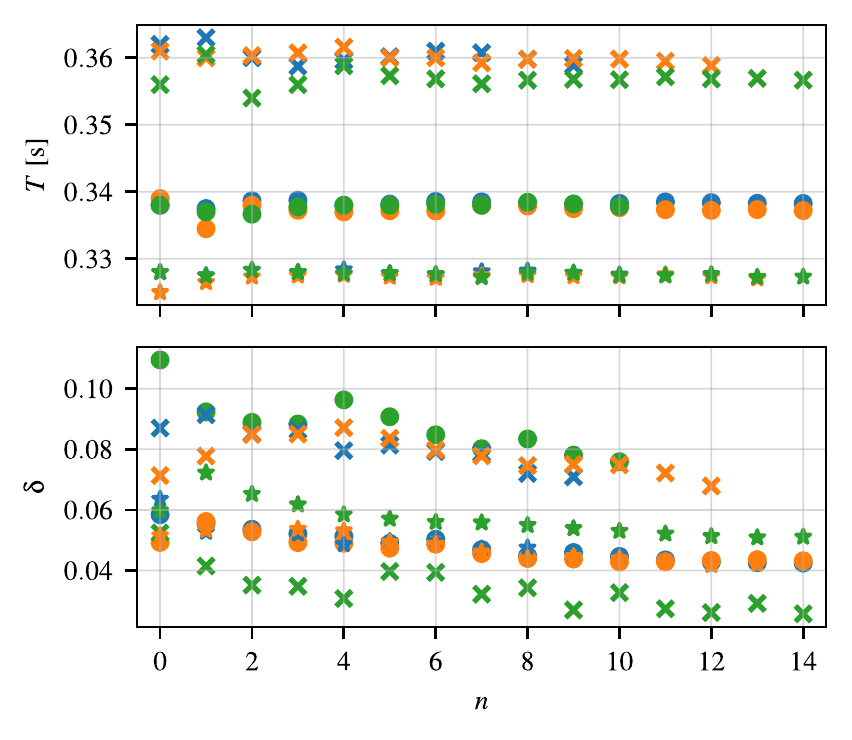}
    \vspace{-0.7cm}
    \caption{Estimated periods of damped oscillations and log decrement for each experiment as a function of the number of peaks $n$. Filled circles $\bullet$ correspond to  $\mathcal{O}_1$, $\times$-s to $\mathcal{O}_2$, and stars $\star$ to $\mathcal{O}_3$; colorwise blue - 0.15m, orange - 0.25m, and green - 0.35m.}
    \label{fig:Ti_deltai}
    \vspace{-0.3cm}
\end{figure}

Fig. \ref{fig:Ti_deltai} shows estimated periods of the damped oscillations and log decrements for each experiment using the data-driven method described in Section \ref{sec:param_est}. Periods of damped oscillations for different orientations are distinctively different and constant. It demonstrates that the orientation affects the natural frequency and that considering only the first natural frequency is sufficient for handling beams with a robot arm.  Unlike $T$, $\delta$ does not clearly depend on the orientation of the beam, indicating that the linear viscous damping model cannot accurately describe the actual damping of the beam. Nonlinear damping models might yield better results but require a more complicated parameter estimation procedure.

For each experiment, we averaged the estimated $\omega_n$ and selected the lowest $\zeta$, as in practice, it is safer to underestimate the damping rather than overestimate. Table \ref{tab:beam_params} shows the estimated beam parameters from data and material properties. The relative errors in $\hat \omega_n$ between analytical and data-driven methods for each orientation are $0.7\ \%$, $1.73\ \%$, and $1.27\ \%$, respectively. Although the error is small, the difference in the equivalent pendulum length $l$ is rather large: $0.52\ \mathrm{m}$ for the analytical model against an average $0.35\ \mathrm{m}$ for the data-driven model. The difference in $\hat l$ between $\mathcal{O}_2$ and  $\mathcal{O}_3$ might be due to additional nonlinearities that the pendulum model cannot explain or estimation errors of the external torque observer.

\begin{table}%[ht]
    \centering
    \caption{Estimated parameters of the beam and pendulum model}
    \label{tab:beam_params}
    \begin{tabular}{ccccc}
        \hline
         method & orientation & $\hat \omega_n$, [rad/s] & $\hat \zeta$, [\%] & $\hat l$, [m] \\
         \hline
        data-driven & $\mathcal{O}_1$ &  18.57 & 0.7 & - \\
        data-driven & $\mathcal{O}_2$ &  17.61 & 0.7 & 0.42 \\
        data-driven & $\mathcal{O}_3$ &  19.19 & 0.7 & 0.28 \\
        analytical & $\mathcal{O}_1$ &  18.44 & - & - \\
        analytical & $\mathcal{O}_2$ &  17.92 & - & 0.52 \\
        analytical & $\mathcal{O}_3$ &  18.95 & - & 0.52 \\
         \hline
    \end{tabular}
\end{table}

% \begin{figure*}[htbp]
\begin{figure*}[]
    \centering
    % \includesvg[keepaspectratio, width=\textwidth]{figures/tau_T123.svg}
    \includegraphics[keepaspectratio, width=\textwidth]{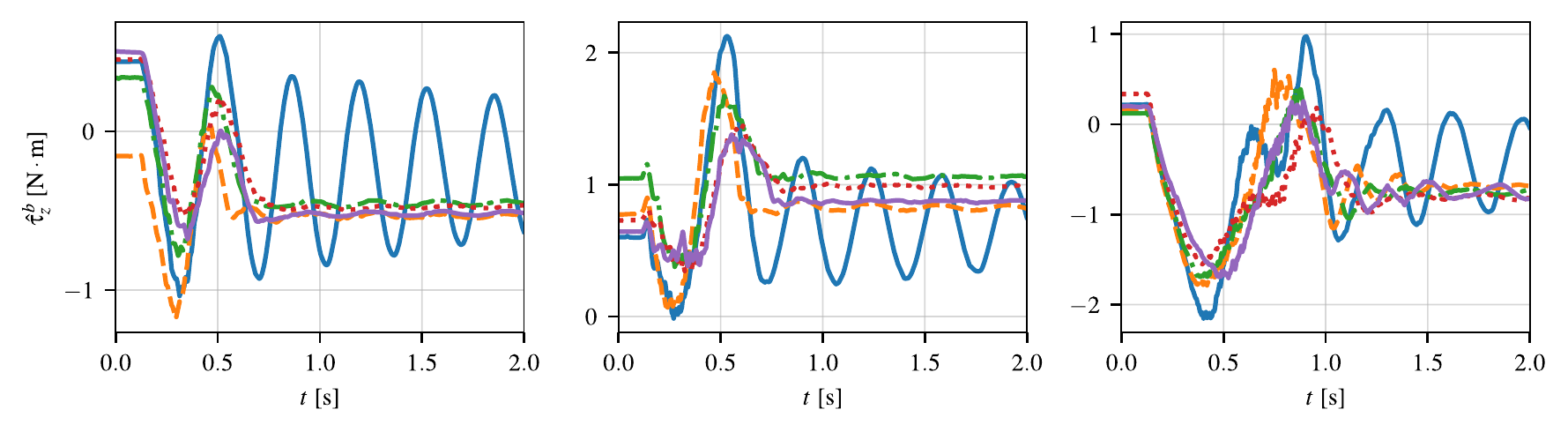}
    \vspace{-0.8cm}
    \caption{External torque estimate $\hat \tau_z^b$ for tasks T1, T2 and T3 from left to right. Solid blue line with large oscillations correspond to a aOCP; solid blue line corresponds to ZV IS. Orange dashed line corresponds to the OCP \eqref{eq:ocp} with the fastest travel time, green dashed-dotted line to the OCP \eqref{eq:ocp} with medium travel time, and, finally, red dotted line corresponds to the OCP \eqref{eq:ocp} with the longest travel time.}
    \label{fig:res_vibr_T123}
\end{figure*}

\subsection{Vibrations suppression}

To validate the proposed approach, we conducted several experiments on our setup, starting from an elementary 1-dimensional task to a complex 3-dimensional task:
\begin{itemize}
    \item [T1] starting from $\bm q_{0} = \bm q_{\mathcal{O}_3}$, as in Fig. \ref{fig:eigen_rot}, move the end-effector by $-0.20\ \mathrm{m}$ along the $X_0-$axis;
    \item [T2] starting from $\bm q_{0} = \bm q_{\mathcal{O}_1}$, as in Fig. \ref{fig:eigen_rot}, move the end-effector by $[0.20\ 0\ -0.20]^\top \ \mathrm{m}$ relative to $\{0\}$;
    \item [T3] starting from $\bm q_{0} = \bm q_{\mathcal{O}_2}$, as in Fig. \ref{fig:eigen_rot}, move the end-effector by $[0.20\ 0\ -0.20]^\top$ $\mathrm{m}$ relative to $\{0\}$ and orient $\{b\}$ alike $\mathcal{O}_1$ but rotated by $90\ \mathrm{deg}$ around $Z_0$.
\end{itemize}
For each task, we compare several feedforward controllers: OCP \eqref{eq:ocp} with three different travel times, $t_{f_1} < t_{f_2} < t_{f_3}$; aOCP with travel time $t_{f_1}$; the zero-vibration (ZV) IS \cite{Singer1990ishaping} with travel time $t_{f_3}$. 

Note that IS does not address trajectory planning; thus, we feed it with the trajectory designed by the aOCP with the shortest feasible travel time. Such optimal input gives IS an advantage in terms of travel time compared with simpler point-to-point planners (e.g., polynomial or trapezoidal trajectory planners) commonly used with IS. 
For T1 and T2, the initial and final orientations of $\{b\}$ coincide, and the natural frequency of the linearized pendulum dynamics does not change along the task (as assumed by ZV IS), making the comparison between OCP and ZV IS reasonably fair. For T3, ZV IS applied to the operational space trajectory yields infeasible trajectories in the joint space after inverse kinematics, unlike T1 and T2; therefore, for T3, we applied ZV IS to the joint space trajectory.

As a performance metric we use the normalized integral of the absolute value of the zero mean residual vibrations (vibrations that persist after the end of the motion)
\begin{align}
    V = \int_{t_f}^{t_f + t_r} \left|\hat \tau_z^b(t) - \bar{\hat{\tau}}_z^b(t)\right|\ dt,
\end{align}
where $\bar{\hat{\tau}}_z^b$ is the average value of $\hat \tau_z^b$ and $t_r$ is a sufficiently long time window such that it contains several of its periods in case of significant vibrations. In this paper, $t_r = 5\ \mathrm{s}$.

Fig. \ref{fig:res_vibr_T123} and  Table \ref{tab:planner_performance} show the performance in terms of residual vibrations of different feedforward controllers for all the tasks.
For T1 and T3, the proposed approach always yields better performance than ZV IS. For T2, OCP outperforms the ZV IS in the case of $t_{f_3}$ but as travel time shortens the performance of OCP decreases falling behind ZV IS.

\begin{table}[]
    \centering
    \caption{Trajectory execution time and performance of used planners}
    \vspace{-0.1cm}
    \label{tab:planner_performance}
    \begin{tabular}{cccccccc}
        \hline \\[-5px]
        \multicolumn{2}{c}{planner} & \multicolumn{3}{c}{$t_f$ [s]} & \multicolumn{3}{c}{$V/V_{\mathrm{aOCP}} \times 100\%$ [\%]}\\
        \hline \\[-5px]
        & & T1 & T2 & T3 & T1 & T2 & T3 \\ \cline{3-5} \cline{6-8}
        aOCP & $t_{f_1}$ & 0.44 & 0.46 & 0.81 & 100 & 100 & 100 \\
        OCP \eqref{eq:ocp} & $t_{f_1}$ & 0.44 & 0.46  & 0.81 & 3.47 & 8.81 & 9.63 \\
        OCP \eqref{eq:ocp} & $t_{f_2}$ & 0.50 & 0.55  & 0.90 & 3.63 & 3.51 & 9.12 \\
        OCP \eqref{eq:ocp} & $t_{f_3}$ & 0.56 & 0.62 & 0.99 & 2.80 & 2.64 & 9.62 \\
        ZV IS & $t_{f_3}$ & 0.56  & 0.62 &  0.99 & 4.47 & 2.92 & 21.57 \\
        \hline
    \end{tabular}
    \vspace{-0.3cm}
\end{table}

Vibration suppression performance of OCP worsens: (i) with the shortening of the travel time, as for T2; (ii) with the growing complexity of the task, as for T3. There are two main reasons for that. First, shorter travel time or greater task complexity require higher accelerations. For such fast trajectories, the internal velocity controller cannot accurately track the reference rendering Assumption \ref{as:arm_model} incorrect. Second, OCP becomes more sensitive to model parameter inaccuracies. The first step toward further reducing residual vibrations should be extending the arm model with the velocity loop model.

Overall, the proposed approach drastically reduces vibrations while being as fast as the fastest feasible aOCP -- which ignores the beam dynamics -- or slightly slower. Moreover, OCP delivers better performance than the well-known ZV IS except for few cases while being much faster (at least 18\%); and allows to trade-off performance and travel time. For some applications, small residual vibrations might not be critical -- e.g., for beam insertion tasks, where the margin in the dimension of the slot, permits some residual vibrations -- while faster task execution time might bring additional economic benefits.

\section{CONCLUSION} \label{sec:disc_conc}
% \subsection{Choice of input shaper}
% In terms of travel time negative input shaper e.g. unity magnitude negative (UM) input shaper [\textbf{?}] outperforms ZV IS. However, if applied to the operational space trajectory UM ZV negative IS returns a trajectory that is infeasible in the joint space. To make the shaped trajectory feasible it is necessary to slow-down the input of the shaper leading to final trajectory with more or less the same travel time as ZV PIS. Moreover, NIS excite higher frequencies making it less attractive overall.

This paper proposes a model-based open-loop method for handling flexible beams with a robot arm without exteroceptive sensors. We approximated the beam with a pendulum and experimentally showed that such approximation is sufficient for describing the dynamics of the beam. For trajectory planning and vibration suppression, we formulated an OCP in the joint space that takes the constraints of the robot into account. The paper also discusses an efficient implementation of the OCP to reduce computational time. Experiments show that the proposed approach outperforms IS in terms of vibration suppression and task execution time for many tasks, and also allows to trade off vibration suppression and task execution time.

Our approach can be extended to include the parameter uncertainty bounds leading to robust feedforward control; for repetitive tasks, iterative learning techniques can be used to further improve performance over iterations. The proposed problem formulation can also be transformed from feedforward to feedback leading to MPC. In this case, the major challenge will be the estimation of the pendulum model's state variables. In an online setting, parameter estimation and control can be combined and solved together using ideas from dual control literature.

\addtolength{\textheight}{-9cm}   % This command serves to balance the column lengths
                                  % so it should come on the page before the last. Mak

%%%%%%%%%%%%%%%%%%%%%%%%%%%%%%%%%%%%%%%%%%%%%%%%%%%%%%%%%%%%%%%%%%%%%%%%%%%%%%%%

%%%%%%%%%%%%%%%%%%%%%%%%%%%%%%%%%%%%%%%%%%%%%%%%%%%%%%%%%%%%%%%%%%%%%%%%%%%%%%%%

%%%%%%%%%%%%%%%%%%%%%%%%%%%%%%%%%%%%%%%%%%%%%%%%%%%%%%%%%%%%%%%%%%%%%%%%%%%%%%%%
% \section*{APPENDIX} \label{sec:appendix}

% \section*{ACKNOWLEDGMENT}

%%%%%%%%%%%%%%%%%%%%%%%%%%%%%%%%%%%%%%%%%%%%%%%%%%%%%%%%%%%%%%%%%%%%%%%%%%%%%%%%

\bibliographystyle{IEEEtran}
\bibliography{IEEEabrv, references}

\end{document}